\pdfoutput=1

\documentclass[pdflatex,sn-mathphys]{sn-jnl}

\usepackage{wrapfig}
\usepackage{graphics}
\usepackage{threeparttable}
\usepackage{amsfonts}
\usepackage{makecell}
\usepackage{nicefrac}
\usepackage{xcolor}
\usepackage{tabularx}
\usepackage{xspace}
\usepackage{amsmath}
\usepackage{graphicx} 
\usepackage{float} 
\usepackage{subfigure} 
\usepackage{threeparttable}
\usepackage{caption}

\jyear{2021}%

\theoremstyle{thmstyleone}%
\theoremstyle{thmstyletwo}%

\theoremstyle{thmstylethree}%

\begin{document}

\title[D-Former for 3D Medical Image Segmentation]{D-Former: A U-shaped Dilated Transformer \\for 3D Medical Image Segmentation}


\author[1]{\fnm{Yixuan} \sur{Wu}}\email{wyx\_chloe@zju.edu.cn}

\author[2]{\fnm{Kuanlun} \sur{Liao}}\email{stevekll@zju.edu.cn}

\author[2]{\fnm{Jintai} \sur{Chen}}\email{jtchen721@gmail.com}

\author[2]{\fnm{Jinhong} \sur{Wang}}\email{wangjinhong@zju.edu.cn}

\author[3]{\fnm{Danny Z.} \sur{Chen}}\email{dchen@nd.edu}

\author*[4,5]{\fnm{Honghao} \sur{Gao}}\email{honghaogao@gachon.ac.kr; gaohonghao@shu.edu.cn}

\author*[6]{\fnm{Jian} \sur{Wu}}\email{wujian2000@zju.edu.cn}

\affil[1]{\orgdiv{School of Medicine}, \orgname{Zhejiang University}, \city{Hangzhou}, \postcode{310030}, \country{China}}

\affil[2]{\orgdiv{College of Computer Science and Technology}, \orgname{Zhejiang University}, \city{Hangzhou}, \postcode{310058}, \country{China}}

\affil[3]{\orgdiv{Department of Computer Science and Engineering}, \orgname{University of Notre Dame}, \city{Notre Dame}, \postcode{IN 46556}, \country{USA}}

\affil[4]{\orgdiv{College of Future Industry}, \orgname{Gachon University}, \city{Seongnam}, \postcode{13120}, \country{Korea}}

\affil[5]{\orgdiv{School of Computer Engineering and Science}, \orgname{Shanghai University}, \city{Shanghai}, \postcode{200444}, \country{China}}

\affil[6]{\orgdiv{Second Affiliated Hospital School of Medicine and School of Public Health}, \orgname{Zhejiang University}, \city{Hangzhou}, \postcode{310030}, \country{China}}


\abstract{Computer-aided medical image segmentation has been applied widely in diagnosis and treatment to obtain clinically useful information of shapes and volumes of target organs and tissues. In the past several years, convolutional neural network (CNN) based methods (e.g., U-Net) have dominated this area, but still suffered from inadequate long-range information capturing. Hence, recent work presented computer vision Transformer variants for medical image segmentation tasks and obtained promising performances. Such Transformers model long-range dependency by computing pair-wise patch relations. However, they incur prohibitive computational costs, especially on 3D medical images (e.g., CT and MRI). In this paper, we propose a new method called Dilated Transformer, which conducts self-attention for pair-wise patch relations captured alternately in local and global scopes. Inspired by dilated convolution kernels, we conduct the global self-attention in a dilated manner, enlarging receptive fields without increasing the patches involved and thus reducing computational costs. Based on this design of Dilated Transformer, we construct a U-shaped encoder-decoder hierarchical architecture called D-Former for 3D medical image segmentation. Experiments on the Synapse and ACDC datasets show that our D-Former model, trained from scratch, outperforms various competitive CNN-based or Transformer-based segmentation models at a low computational cost without time-consuming per-training process.}




\maketitle

\section{Introduction}\label{sec:introduction}

Medical image segmentation, as one of the critical computer-aided medical image analysis problems, aims to capture precisely the shapes and volumes of target organs and tissues by pixel-wise classification, obtaining clinically useful information for diagnosis, treatment, and intervention. With the recent development of deep learning methods and computer vision algorithms, medical image segmentation has been revolutionized and achieved remarkable progresses (e.g., automatic liver and tumor lesion segmentation~\cite{autoliverseg}, brain tumor segmentation~\cite{brainseg}, and multiple sclerosis (MS) lesion segmentation~\cite{lesionseg}).

Fully convolutional network (FCN)~\cite{fcnn} was first proved effective for general image segmentation tasks, which became a predominant technique for medical image segmentation~\cite{fcn1,fcn2,fcn3,fcn4,fcn5}. However, it was observed that vital details can be missing with the decrease of feature map sizes when FCN models go deeper. To this end, a family of U-shaped networks~\cite{unet,unet1,unet2,unet3,unet4,unet5,unet6,vnet} extended the sequential FCN frameworks into encoder-decoder type architectures, alleviating the spatial information loss using skip-connections. Competitive solutions were given by DeepLab models~\cite{deeplab0,deeplab,deeplab2,deeplab3}, which applied atrous convolutions instead of pool layers to expand the receptive field and introduced fully connected conditional random field (CRF) to maintain fine details. Although these CNN-based methods have achieved great performances in medical image segmentation tasks, they still suffered from limited receptive fields and cannot capture long-range dependency, leading to sub-optimal accuracy and failing to meet the needs of various medical image segmentation scenarios.



Inspired by the success of Transformer with its self-attention mechanism in natural language processing (NLP) tasks~\cite{transformer,bert}, researchers tried to adapt Transformers~\cite{vit,deit,detr,deformabledetr} to computer vision in order to compensate the locality of CNNs. The self-attention mechanism in Transformers is able to compute pair-wise relations between patches globally, consequently achieving feature interactions across a long range. Non-local neural network~\cite{nonlocal} was the first to adopt self-attention mechanism to complement CNNs for modeling pixel-level long-range dependency for visual recognition tasks. Then, the Vision Transformer (ViT)~\cite{vit} utilized a pure Transformer framework to deal with vision tasks, treating an image as a collection of spatial patches. Recently, Transformers have achieved excellent outcomes on a variety of vision tasks~\cite{swintransformer,pvt,ant,nnformer,detection1,detection2}, including image recognition~\cite{swintransformer,pvt,ant,t2tvit,volo}, semantic segmentation~\cite{nnformer}, and object detection~\cite{detection1,detection2}. On semantic medical image segmentation, Transformer-combined architectures can be divided into two categories: the main one adopts self-attention like operations to complement  CNNs~\cite{transunet,unetr,transfuse,cotr}; the other uses pure Transformers to constitute encoder-decoder architectures so as to capture deep representations and predict the class of each image pixel~\cite{swinunet,dstransunet,nnformer,missformer}.

Although the above medical image segmentation methods were promising and yielded good performance to some extent, they still suffered considerable drawbacks. (1) The majority of these Transformer segmentation models were designed for 2D images~\cite{transunet,transfuse,swinunet,dstransunet,missformer}. For 3D medical images (e.g., 3D MRI scans), they may divide the input images into 2D slices and process the individual slices with the 2D models, which could lose useful 3D contextual information. (2) Compared with common 2D natural scene images, processing 3D medical images inevitably incurs larger model sizes and computation costs, especially when computing global feature interactions with self-attention in vanilla Transformer~\cite{transformer} (see more details in Sec.~\ref{sec:Dblock}). Although some adaptations were proposed to reduce the operation scopes of self-attention~\cite{xcit,swintransformer,pvt,ant,litetransformer,define,delight} (e.g., Pyramid Vision Transformer~\cite{pvt} used progressive scaling pyramids to reduce the computation of large feature maps), they could lead to insufficient global information fusion.
(3) The self-attention operation in Transformers was shown to be permutation-equivalence~\cite{transformer}, which omits the order of patches in an input sequence. However, the permutation-equivalence nature can be detrimental to medical image segmentation since segmentation results are often highly position-correlated. Prior work usually utilized absolute position encoding (APE)~\cite{transformer} or relative position encoding (RPE)~\cite{rpe,swintransformer} to supplement position information. But, APE requires a pre-given and fixed patch amount and thus fails to generalize to different image sizes, while RPE ignores the absolute position information that can be a vital cue in medical images (e.g., the positions of bones are often relatively stable).


To address the drawbacks above, we propose a new efficient model called \textbf{Dilated Transformer} (D-Former) to directly process 3D medical images (instead of dealing with 2D slices of 3D images independently) and predict volumetric segmentation masks. Our proposed D-Former is a 3D U-shaped architecture with hierarchical layers, and employs skip connections from encoder to decoder following~\cite{unet,unet1,unet2,unet3,unet4,unet5,unet6,vnet}. This model's stem is constructed with eight D-Former blocks, each of which consists of several local scope modules (LSMs) and global scope modules (GSMs). The LSM conducts self-attention locally, focusing on fine information capturing. The GSM performs global self-attention on uniformly sampled patches, aiming to explore rough and long-range dependent information at low cost. The LSMs and GSMs are arranged in an alternate manner to achieve local and global information interaction. For drawback (3) above, we manage to incorporate position information among patches in a more dynamic manner. Inspired by~\cite{cpe,xcit}, we utilize depth-wise convolutions~\cite{dwconv} to learn position information, which can help provide useful position cues in medical image segmentation.



Benefiting from these designs, our proposed D-Former model could be more suitable for medical image segmentation tasks and yield better segmentation accuracy. The main contributions of this work are as follows:
\begin{itemize}
  \item [(1)] 
  We construct a 3D Transformer based architecture which allows to process volumetric medical images as a whole and thus spatial information along the depth dimension of 3D medical images to be fully captured.      
  \item [(2)]
  We design local scope modules (LSMs) and global scope modules (GSMs) to increase the scopes of information interactions without increasing the patches involved in computing self-attention, which helps reduce computational costs.
  \item [(3)]
   To further incorporate relative and absolute position information among patches, we apply a dynamic position encoding method to learn it from the input directly. As a result, an inherent problem of common Transformers, permutation-equivalence \cite{transformer}, could be considerably alleviated. 
   \item [(4)]
   Extensive experimental evaluations show that our model outperforms state-of-the-art segmentation methods in different domains (e.g., CT and MRI), with smaller model sizes and less FLOPs than the known methods. 
\end{itemize}

\section{Related Work}


 

\subsection{CNN-based Segmentation Networks}
Since the advent of the seminal U-Net model~\cite{unet}, many CNN-based networks have been developed~\cite{vnet,resunet,raunet,nnunet,denseunet}. As for the design of skip connections, U-Net++~\cite{unet++} and U-Net3+~\cite{unet3+} were proposed to attain dense connections between encoder and decoder. In addition, regarding the locality of CNNs, researchers have designed different kinds of mechanisms to enlarge the receptive field, such as larger kernel~\cite{largekernel}, dilated convolution module~\cite{dilated2,dilated3}, pyramid pooling module~\cite{pyramid2,pyramid3}, and deformable convolution module~\cite{deformable2,deformable3}. In particular, dilated convolution is an ingenious design, which expands the convolution kernel by inserting holes between its consecutive elements. This design has been adopted by various segmentation models, achieving good performance compared with the original convolution-based methods. Our Dilated Transformer also obtains a key idea from this design and aims to conduct self-attention in a patch skipping manner (see Sec.~\ref{sec:Dblock} for details).

\subsection{Visual Transformer Variants}
Transformer and its self-attention mechanism were first designed for sequence modeling and transduction tasks in the domain of natural language processing (NLP), achieving state-of-the-art performance~\cite{transformer,bert}. Inspired by tremendous success of Transformer in NLP, researchers 
adapted Transformer to computer vision tasks. The first attempt was vision Transformer (ViT)~\cite{vit}, which needed huge pre-training datasets. In order to overcome this weakness, DeiT~\cite{deit} was proposed, and this work provided a wide range of training strategies with knowledge distillation, which contributed to better performance of vanilla Transformer. There were different kinds of adaptations for vanilla Transformer, such as Swin Transformer~\cite{swintransformer}, pyramid vision Transformer~\cite{pvt}, Transformer in Transformer~\cite{tint}, and aggregating nested Transformers~\cite{ant}. In particular, Swin Transformer showed great success in various computer vision tasks with its elegant shift window mechanism and hierarchical architecture. Our proposed D-Former is inspired by Swin Transformer's local-global combining scopes of information interactions.

\subsection{Transformers for Segmentation Tasks}
As mentioned above, Transformers used in medical image segmentation methods can be divided into two categories. In the main category, Transformer and its self-attention mechanism are utilized as a supplement for the convolution-based stem. SETR~\cite{setr} was proposed to apply Transformer as encoder to extract features for segmentation tasks. In medical imaging, many models with Transformers focused on segmentation tasks. In TransUNet~\cite{transunet}, convolutional layer was used as a feature extractor to obtain detailed information from raw images; it then generated feature maps which are put into Transformer layer to obtain global information. UNETR~\cite{unetr} proposed a 3D Transformer-combining architecture for medical images, which treated Transformer layer as encoder to extract features and convolutional layer as decoder. A great amount of such work focused on taking advantage of both Transformer's long-range dependency and CNN's inductive bias. In the other category, Transformer is regarded as the main stem for building the whole architecture~\cite{swinunet,dstransunet,nnformer,missformer}. MedT~\cite{medt} proposed the Gated Axial-Attention mechanism and used Gated Axial Transformer layer to build the whole architecture. Swin-Unet~\cite{swinunet} used the Swin Transformer block as a basic unit to build a U-shape architecture with skip connections. DS-TransUNet~\cite{dstransunet} further extended Swin-Unet by adding another encoder pathway for input of different sizes. nnFormer~\cite{nnformer} modified the up-sampling and down-sampling blocks with convolutions based on the Swin-Unet architecture. 

\begin{figure*}
\centering 
\includegraphics[width=1.0\textwidth]{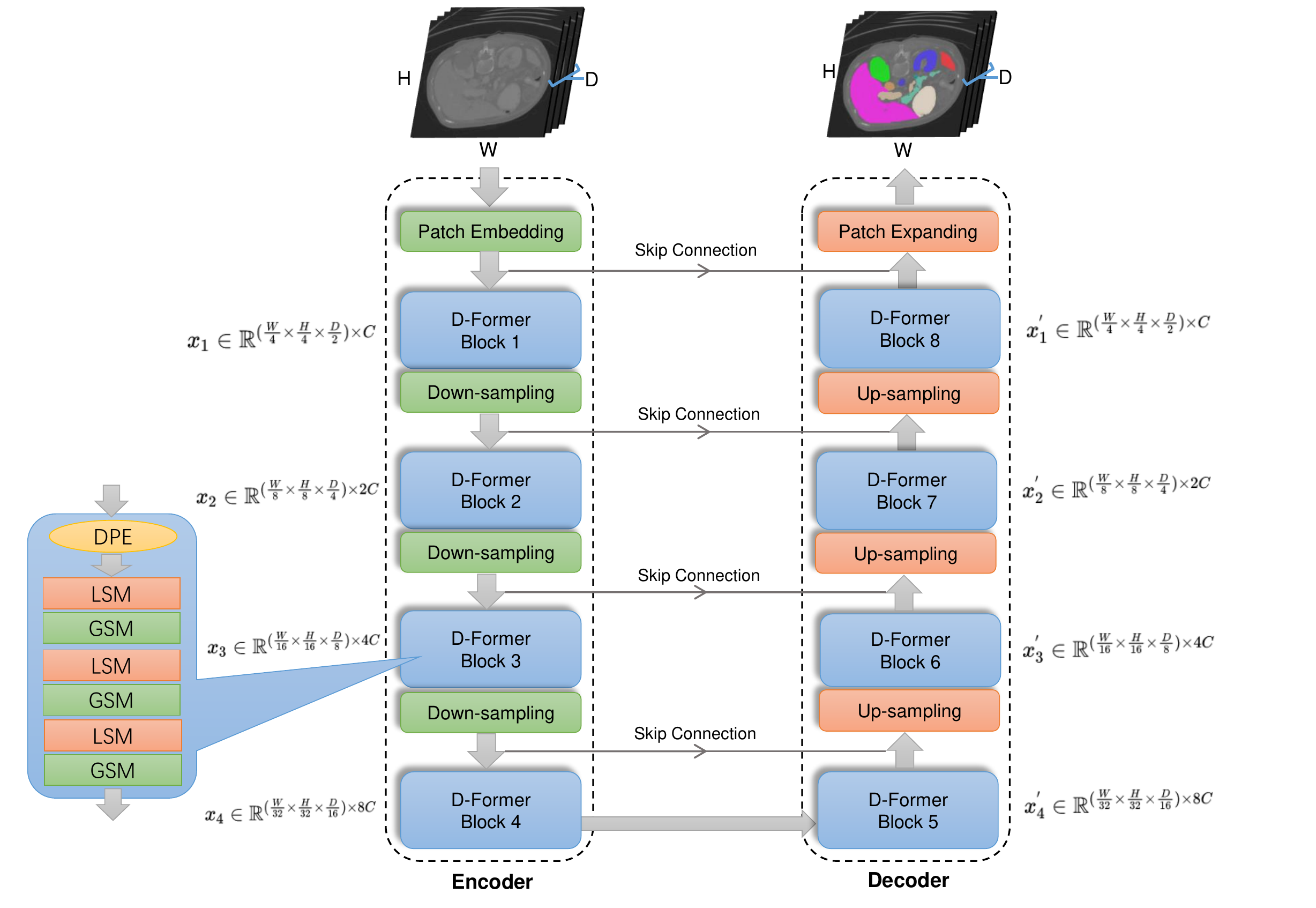}
\caption{The overall architecture of our D-Former model. Each D-Former block is constructed with one dynamic position encoding block (DPE) and several local scope modules (LSMs) and global scope modules (GSMs). The input size of the D-Former block $i$ is reported sideward, and the output sizes are the same as the corresponding input sizes. The values in round brackets denote the numbers of patches, which are regarded as one dimension when computed in Transformers (i.e., $(\frac{W}{4} \times \frac{H}{4} \times \frac{D}{2})$, $(\frac{W}{8} \times \frac{H}{8} \times \frac{D}{4})$, $(\frac{W}{16} \times \frac{H}{16} \times \frac{D}{8})$, $(\frac{W}{32} \times \frac{H}{32} \times \frac{D}{16})$).} 
\label{fig:architecture} 
\end{figure*}

\section{Method}
\subsection{The Overall Architecture}
Our proposed D-Former model is outlined in Fig.~\ref{fig:architecture}, which is a hierarchical encoder-decoder architecture. The encoder pathway consists of one patch embedding layer for transforming 3D images into sequences and four proposed D-Former blocks for feature extraction with three down-sampling layers in between them. The first, second, and fourth D-Former blocks each consist of one local scope module (LSM) and one global scope module (GSM) respectively, while the third D-Former block has three LSMs and three GSMs, in which the LSMs and GSMs are arranged in an alternate manner. The decoder pathway is symmetric to the encoder pathway, which also has four D-Former blocks, three up-sampling layers, and one patch expanding layer. In addition, skip connections are used to transfer information from the encoder to the decoder at the corresponding levels. The feature maps from the encoder are concatenated with the corresponding feature maps along the channel dimension, which may compensate for the loss of fine-grained information as the model goes deep.

In this section below, we will present the components of \textbf{D-Former} one by one, including the patch embedding and patch expending layers (Sec.~\ref{sec:pepe}), the D-Former block and its major modules, the local scope module and global scope module (Sec.~\ref{sec:Dblock}), the down-sampling and up-sampling operations (Sec.~\ref{sec:dsus}), and the dynamic position encoding block (Sec.~\ref{sec:dpe}).

\subsection{Patch Embedding \& Patch Expending}\label{sec:pepe}
Similar to common Transformers in computer vision, after data augmentation, an input 3D medical image
$x \in \mathbb{R}^{W \times H \times D}$
first goes through a patch embedding layer and is divided into a series of patches of size $4\times4\times2$ each, and then is projected into $C$ channel dimensions by linear projection to yield a feature map (denoted by $x_1$) of size
$(\frac{W}{4} \times \frac{H}{4} \times \frac{D}{2}) \times C$, where $(\frac{W}{4} \times \frac{H}{4} \times \frac{D}{2})$ denotes the number of patches and $C$ is the number of the channel dimensions. Hence, the input 3D image is reorganized as a sequence (of length $(\frac{W}{4} \times \frac{H}{4} \times \frac{D}{2})$) and can be directly fed to a Transformer architecture. The final patch expending layer is used to restore the feature map to the original input size, and utilizes a segmentation head (like 3D UNet~\cite{3dunet}) to yield pixel-wise segmentation masks.


\subsection{D-Former Blocks}\label{sec:Dblock}
After patch embedding, $x_{1}$ is directly fed to D-Former block 1. In the processing by Transformer block 1, $x_1$ is first processed by a new dynamic position encoding block that embeds position information into feature maps (see details in Sec.~\ref{sec:dpe}), and then it is operated by the Local Scope Module (LSM) and Global Scope Module (GSM) alternatively to extract higher-level features. The other D-Former Blocks process the corresponding input features similarly, and the feature map sizes are provided in Fig.~\ref{fig:architecture}.

\begin{figure}
\centering 
\includegraphics[width=0.5\textwidth]{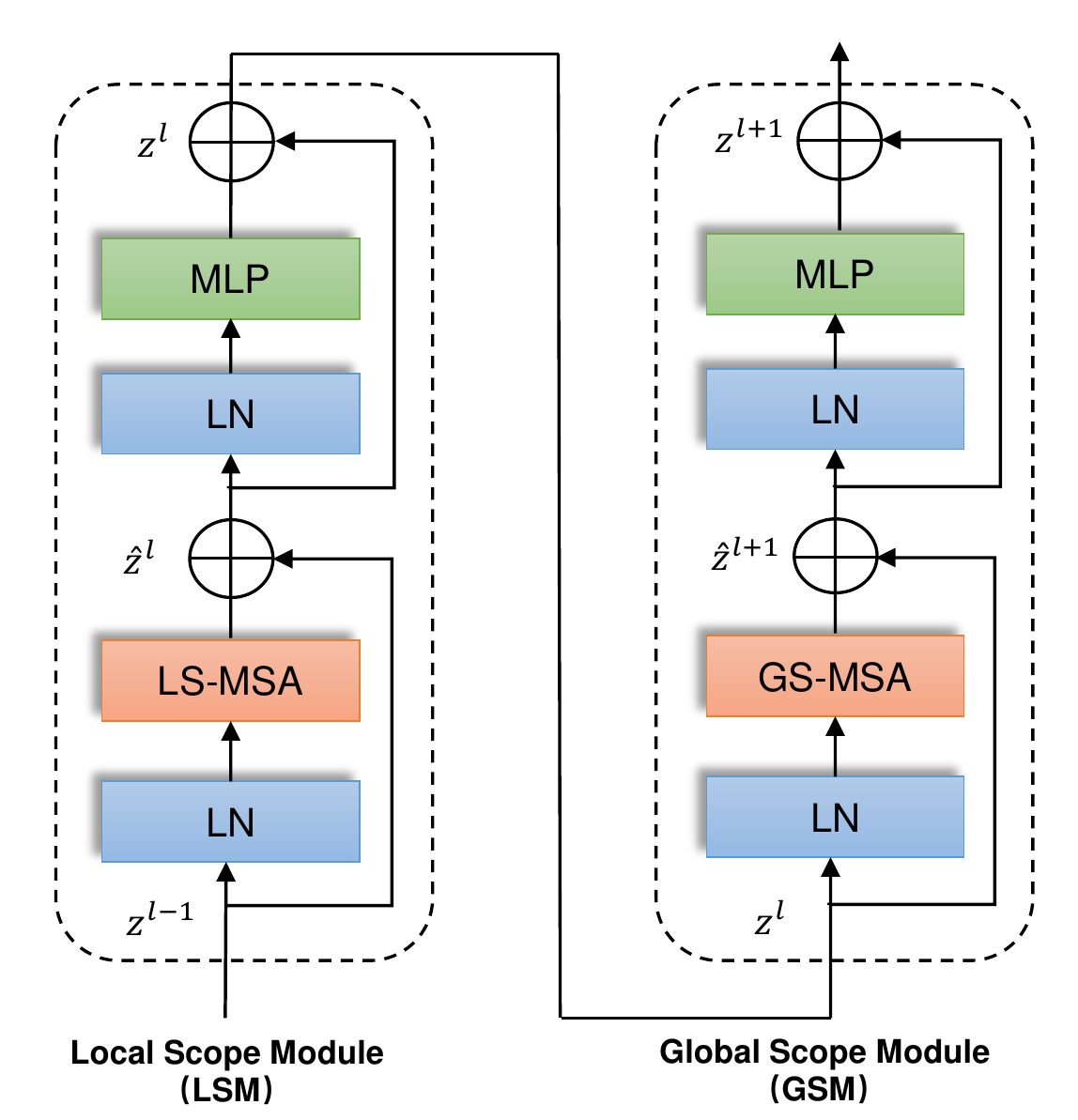}
\caption{The local scope module (LSM) and global scope module (GSM), which should be arranged in pair to combine local and global information. } 
\label{fig:transformer} 
\end{figure}

\subsubsection{Local Scope Module and Global Scope Module}
The local scope module (LSM) and global scope module (GSM) are presented to capture local and global features respectively, which employ two different self-attention operations, called local scope multi-head self-attention (LS-MSA) and global scope multi-head self-attention (GS-MSA).
As shown in Fig.~\ref{fig:transformer}, an LSM is composed of a LayerNorm layer~\cite{ln}, a proposed LS-MSA, another LayerNorm layer, and a Multilayer Perceptron (MLP), in sequence, with two residual connections to prevent gradient vanishing~\cite{transformer}. In a GSM, the LS-MSA is replaced by a proposed GS-MSA, and the other components are kept the same as the LSM. To allow local features and global features to be captured and fused well, we arrange the LSM and GSM alternatively in each D-Former block. With these components, their operations are formally defined as:
%
\begin{gather}\label{eq:RS}
    \hat{z}^{l}=\textit{LS-MSA}\left(\textit{LN}\left(z^{l-1}\right)\right)+z^{l-1}, \\
    z^{l}=\textit{MLP}\left(\textit{LN}\left(\hat{z}^{l}\right)\right)+\hat{z}^{l}, \\
    \hat{z}^{l+1}=\textit{GS-MSA}\left(\textit{LN}\left(z^{l}\right)\right)+z^{l}, \\
    z^{l+1}=\textit{MLP}\left(\textit{LN}\left(\hat{z}^{l+1}\right)\right)+\hat{z}^{l+1},
\end{gather}
where $\hat{z}^{l}$ and $z^{l}$ denote the outputs of LS-MSA and the corresponding MLP respectively, and $\hat{z}^{l+1}$ and $z^{l+1}$ denote the outputs of GS-MSA and the corresponding MLP respectively.

\begin{figure*}
\centering  
\subfigure[LS-MSA]{
\label{fig:lsmsa}
\includegraphics[width=0.4\textwidth]{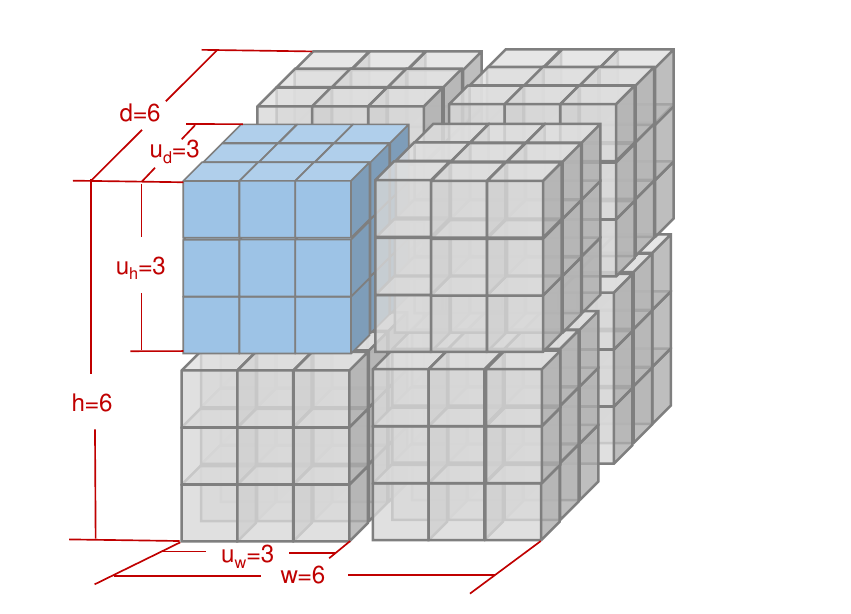}}
\subfigure[GS-MSA]{
\label{fig:gsmsa}
\includegraphics[width=0.41\textwidth]{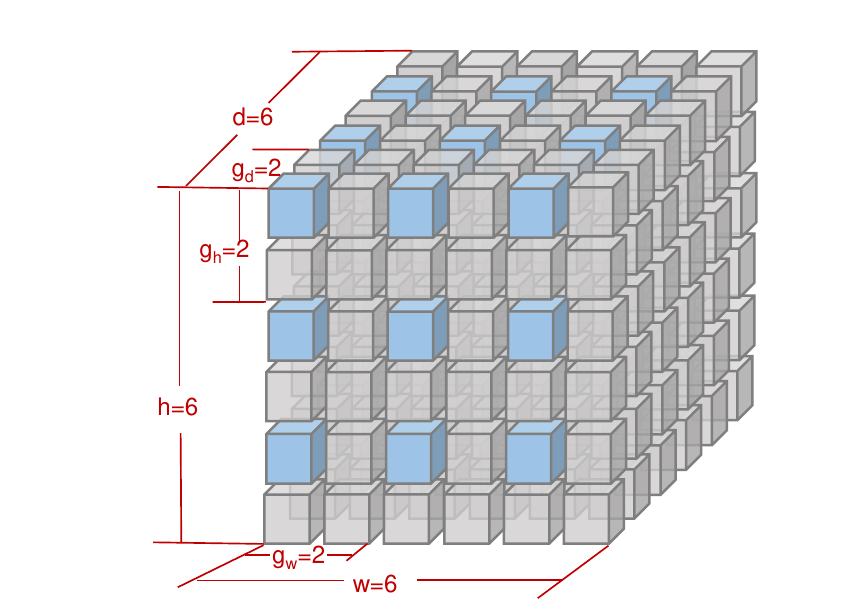}}
\caption{(a) The local scope multi-head self-attention: The self-attention is conducted in a local unit (colored in blue) where the patches are adjacent. (b) The global scope multi-head self-attention: The self-attention is conducted in a global unit (colored in blue) where patches are picked every $g$-th patch across the feature map. A small cube represents one patch. The feature map size is set as $6\times 6\times 6$ and the unit size is $3\times 3\times 3$ as an example. We color only the patches of one unit in blue for illustration; the other grey patches are also utilized to construct seven other units in both LS-MSA and GS-MSA.}
\label{fig:attention}
\end{figure*}

\subsubsection{Local Scope Multi-head Self-Attention (LS-MSA)}
A vanilla Transformer conducts self-attention in a global scope in order to capture pair-wise relationships between patches, leading to quadratic complexity with respect to the number of patches. However, due to the fact that 3D medical images would increase computation inevitably, this original self-attention would not be suitable for 3D medical image related tasks, especially for semantic segmentation with dense prediction targets.
Under such circumstances, as illustrated in Fig.~\ref{fig:lsmsa}, we first evenly divide a whole feature map into non-overlapping units (we denote the number of patches in each unit by $u_{d} \times u_{h} \times u_{w}$, where $u_d$ denotes the number of 
patches in one unit along the depth dimension $D$, $u_h$ along the height dimension $H$, and $u_w$ along the width dimension $W$), and conduct self-attention within each unit. In this way, the computational complexity will be reduced to linear in terms of the number of patches in the whole feature map. The computational complexity ($\Omega$) of these two different self-attention mechanisms are computed as: 
\begin{gather}\label{eq:complexity}
    \Omega(\textit{MSA})=4 d h w C^{2}+2(d h w)^{2} C, \\
    \Omega(\textit{LS-MSA})=4 d h w C^{2}+2 u_{d} u_{h} u_{w} d h w C,
\end{gather}
where $u_{d} u_{h} u_{w}$ denotes the number of patches in one unit, and $dhw$ denotes the number of patches in the whole feature map ($d$, $h$, and $w$ denote the depth, height, and width of the feature map, respectively). In most cases, $u_{d} u_{h} u_{w}\ll dhw$. The \textit{Softmax} operation is omitted when computing the computational complexity.

\subsubsection{Global Scope Multi-head Self-Attention (GS-MSA)}
The LS-MSA performs self-attention only within each local unit, which lacks global information interaction and long-rang dependency. To address this issue, we design a global scope multi-head self-attention mechanism to attain information interaction across different units in a dilated manner. As illustrated in Fig.~\ref{fig:gsmsa}, for a whole feature map, we pick one patch every g distance along each dimension and form a unit with all the patches thus picked, on which self-attention would then be conducted. Likewise, we pick the other patches to form new units, until all the patches are utilized. Hence, the receptive field in computing self-attention will be enlarged but the number of patches involved will not be increased, which means that it would not increase the computational cost while getting access to long-range information interaction. To keep consistency between LSM and GSM, we set 
$d=g_{d} \times u_{d}$,
$h=g_{h} \times u_{h}$,
and $w=g_{w} \times u_{w}$,
which ensure that the numbers of units in LSM and GSM are kept the same. Here, $d \times h \times w$ denotes the number of patches in the whole feature map, $u_{d} \times u_{h} \times u_{w}$ denotes the number of patches in one unit, and $g_{d}$, $g_{h}$, and $g_{w}$ denote the distance between two nearest patches picked along the depth dimension $D$, height dimension $H$, and width dimension $W$, respectively.

\subsection{Down-sampling \& Up-sampling}\label{sec:dsus}
Between every two adjacent D-Former blocks of the encoder, a down-sampling layer is utilized to merge patches for further feature fusion. Specifically, a down-sampling layer concatenates the feature maps of $2\times 2\times 2$ neighboring patches (2 neighboring patches along the width, height, and depth dimensions, respectively), reducing the number of patches by 8 times. Then, a fully connected layer is utilized to reduce the feature channel size by 4 times to ensure that the channel size can be doubled after each down-sampling layer. Thus, the output feature maps of each down-sampling layer will be
$x_{2} \in \mathbb{R}^{(\frac{W}{8} \times \frac{H}{8} \times \frac{D}{4}) \times 2C}$,
$x_{3} \in \mathbb{R}^{(\frac{W}{16} \times \frac{H}{16} \times \frac{D}{8}) \times 4C}$, and
$x_{4} \in \mathbb{R}^{(\frac{W}{32} \times \frac{H}{32} \times \frac{D}{16}) \times 8C}$, respectively.
In reverse to the down-sampling layers, four up-sampling layers of the decoder are used to enlarge the low-resolution feature maps and reduce the number $C$ of the channel dimensions. In this way, our model will be able to extract features in a multi-scale manner and yield better segmentation accuracy.

\subsection{The Dynamic Position Encoding Block}\label{sec:dpe}
The depth-wise convolution (\textit{DW-Conv}) is a type of convolution that applies a single convolutional filter for each input channel instead of for all channels as in a common convolution, which can decrease the computational cost. We apply 3D depth-wise convolution~\cite{dwconv} to the input feature maps (or images) once in every D-Former block to learn position information. Then the learned position information will be added to the original input $x_i$ as:
\begin{gather}\label{eq:DW}
    x_i^\prime = \textit{Resize}(\textit{DW-Conv}(\textit{Resize}(x_i)))+x_i,
\end{gather}
where $x_i$ denotes the input feature maps of the $i$-th D-Former block and $x_i^\prime$ denotes the output feature maps embedded with position information. \textit{Resize} is used to adjust the dimensions of feature maps $x_i$ to cater the input need of DW-Convolution.

In this way, position information among patches can be extracted by a DW-Convolution. Given the fact that position information could be dynamically learned based on the input $x$ itself, a drawback in the previous work that requires a fixed number of patches can be avoided. In addition, the convolution’s inherent nature of translation-invariance can be utilized to increase the stability and generalization performance~\cite{transinvar}.

\section{Experiments}
\subsection{Datasets}

The \textbf{Synapse multi-organ segmentation (Synapse)} dataset includes 30 axial contrast-enhanced abdominal CT scans. Following the training-test split in~\cite{transunet}, 18 of the 30 scans are used for training and the remaining ones are for testing. We take the average Dice Similarity Coefficient (DSC)~\cite{vnet} as the measure for evaluating the segmentation performances of eight target organs, including aorta, gallbladder, kidney (L), kidney (R), liver, pancreas, spleen, and stomach.

The \textbf{Automated Cardiac Diagnosis Challenge (ACDC)} dataset contains 150 magnetic resonance imaging (MRI) 3D cases collected from different patients, and each case covers a heart organ from the base to the apex of the left ventricle. Following the setting in~\cite{transunet}, only 100 well-annotated cases are used in the experiments, and the training, validation, and test data are partitioned with the ratio of 7: 1: 2. For fair comparison, the average DSC is employed to evaluate the segmentation performances following the previous work~\cite{transunet}, and three key parts of the heart are chosen, including the right ventricle (RV), myocardium (Myo), and left ventricle (LV).

\subsection{Implementation Setup}
\noindent \textbf{Pre-training.} Our D-Former model is trained from scratch, which means that we initialize the model's weights randomly. Note that in common practice, pre-training is important to Transformer-based models. This is because the pre-training process provides generalized representations and prior knowledge for down-stream tasks. For example, in vision Transformer (VIT)~\cite{vit}, it considered that the model performance depends heavily on pre-training, and its experiments verified this view. Besides, lots of known medical image segmentation methods used pre-trained weights to initialize their models~\cite{nnformer,transunet,transbts,transfuse,levitunet}. However, the pre-training process of Transformer-based models brings up two issues. First, the pre-training process usually incurs high computational complexity in terms of time or computation consumed. Second, for medical images, there are few complete and acknowledged sizable datasets for pre-training (in comparison, ImageNet \cite{imagenet} is available for natural scene images), and the domain gap between natural images and medical images makes it hard for medical image segmentation models to use existing large natural image datasets directly. For these reasons, we choose to train our D-Former model from scratch, which nevertheless yields promising performance that surpasses state-of-the-art methods with pre-training.\\

\noindent \textbf{Implementation details.}
Our proposed D-Former is implemented on PyTorch 1.8.0, and all the experiments are trained on an NVIDIA GeForce RTX 3090 GPU with 24 GB memory. The batch size during training is 2 and during inference in 1. The SGD optimizer~\cite{sgd} with momentum 0.99 is used. The initial learning rate is 0.01 with weight decay of 3e-5. The poly learning rate strategy~\cite{polylr} is utilized with the maximum training epochs of 3000 for the Synapse dataset and 1500 for the ACDC dataset. \\

\noindent \textbf{Loss function.} The cross entropy loss and Dice loss are both widely used for general segmentation tasks. However, since the cross entropy loss is apt to perform well for uniform class distribution while Dice loss is more suitable for target objects of large sizes~\cite{losssurvey}, each of them alone may not be effective for medical image segmentation tasks that involve imbalanced classes and target objects of small sizes. Thus, our loss function combines the binary cross entropy loss $Y$~\cite{celoss} and Dice loss $\hat{Y}$~\cite{vnet} together, which is defined as:
\begin{gather}\label{eq:loss}
    \mathcal{L}(Y, \hat{Y})=-\frac{1}{N} \sum_{n=1}^{N}\left(\frac{1}{2} \cdot Y_{n} \cdot \log \hat{Y}_{n}+\frac{2 \cdot Y_{n} \cdot \hat{Y}_{n}}{Y_{n}+\hat{Y}_{n}}\right)
\end{gather}
where $Y_{n}$ and $\hat{Y}_{n}$ denote the ground truth and predicted probabilities of the $n^{th}$ image respectively, and $N$ is the batch size. 


\begin{table}[t]
\centering
\caption{Segmentation performances of different methods on the Synapse dataset (average Dice  Similarity Coefficient (DSC) in \%). }
\label{table:synapse}
\begin{threeparttable}
\setlength{\tabcolsep}{1mm}
\resizebox{\textwidth}{26mm}{
\begin{tabular}{l|c|cccccccc}
\hline
\multicolumn{1}{c|}{Method}         & Average        & Aotra          & Gallbladder    & Kidnery (L)     & Kidnery (R)     & Liver          & Pancreas       & Spleen         & Stomach        \\ \hline
V-Net           & 68.81          & 75.34          & 51.87          & 77.10          & 80.75          & 87.84          & 40.04          & 80.56          & 56.98          \\
DARR            & 69.77          & 74.74          & 53.77          & 72.31          & 73.24          & 94.08          & 54.18          & 89.90          & 45.96          \\
R50 U-Net       & 74.68          & 87.74          & 63.66          & 80.60          & 78.19          & 93.74          & 56.90          & 85.87          & 74.16          \\
R50 Att-UNet    & 75.57          & 55.92          & 63.91          & 79.20          & 72.71          & 93.56          & 49.37          & 87.19          & 74.95          \\
U-Net           & 76.85          & 89.07          & 69.72          & 77.77          & 68.60          & 93.43          & 53.98          & 86.67          & 75.58          \\
Att-UNet        & 77.77          & 89.55          & 68.88          & 77.98          & 71.11          & 93.57          & 58.04          & 87.30          & 75.75          \\
VIT             & 67.86          & 70.19          & 45.10          & 74.70          & 67.40          & 91.32          & 42.00          & 81.75          & 70.44          \\
R50 VIT         & 71.29          & 73.73          & 55.13          & 75.80          & 72.20          & 91.51          & 45.99          & 81.99          & 73.95          \\
TransUNet       & 77.48          & 87.23          & 63.13          & 81.87          & 77.02          & 94.08          & 55.86          & 85.08          & 75.62          \\
Swin-UNet       & 79.13          & 85.47          & 66.53          & 83.28          & 79.61          & 94.29          & 56.58          & 90.66          & 76.60          \\
TransClaw U-Net & 78.09          & 85.87          & 61.38          & 84.83          & 79.36          & 94.28          & 57.65          & 87.74          & 73.55          \\
LeVit-Unet-384  & 78.53          & 87.33          & 62.23          & 84.61          & 80.25          & 93.11          & 59.07          & 88.86          & 72.76          \\
nnFormer        & 87.40          & 92.04          & 71.09          & 87.64          & 87.34          & 96.53          & \textbf{82.49} & 92.91          & \textbf{89.17} \\
MISSFormer      & 81.96          & 86.99          & 68.65          & 85.21          & 82.00          & 94.41          & 65.67          & 91.92          & 80.81          \\
D-Former  & \textbf{88.83} & \textbf{92.12} & \textbf{80.09} & \textbf{92.60} & \textbf{91.91} & \textbf{96.99} & 76.67          & \textbf{93.78} & 86.44          \\ \hline
\end{tabular}}
\footnotesize               
* The DSC values of the compared models are from open source of the original papers. 
\end{threeparttable}
\end{table}

\begin{table}[t]
\centering
\caption{Segmentation performances of different methods on the ACDC dataset (average Dice  Similarity Coefficient (DSC) in \%).}
\label{table:acdc}
\setlength{\tabcolsep}{4mm}
\begin{threeparttable}
\begin{tabular}{l|c|ccc}
\hline
\multicolumn{1}{c|}{Method}       & Average        & RV             & Myo            & LV             \\ \hline
R50 U-Net      & 87.55          & 87.10          & 80.63          & 94.92          \\
R50 Att-UNet   & 86.75          & 87.58          & 79.20          & 93.47          \\
VIT            & 81.45          & 81.46          & 70.71          & 92.18          \\
R50 VIT        & 87.57          & 86.07          & 81.88          & 94.75          \\
TransUNet      & 89.71          & 88.86          & 84.54          & 95.73          \\
Swin-UNet      & 90.00          & 88.55          & 85.62          & 95.83          \\
LeVit-Unet-384 & 90.32          & 89.55          & 87.64          & 93.76          \\
nnFormer       & 91.78          & 90.22          & 89.53          & 95.59          \\
MISSFormer     & 87.90          & 86.36          & 85.75          & 91.59          \\
D-Former & \textbf{92.29} & \textbf{91.33} & \textbf{89.60} & \textbf{95.93} \\ \hline
\end{tabular}
\footnotesize               
* The DSC values of the compared models are from open source of the original papers. 
\end{threeparttable}
\end{table}

\subsection{Quantitative Results}
We evaluate the performance of our proposed D-Former model on the Synapse and ACDC datasets, and compare with various state-of-the-art models, including V-Net~\cite{vnet}, DARR~\cite{darr}, R50 U-Net~\cite{unet}, R50 Att-UNet~\cite{attunet}, U-Net~\cite{unet}, Att-UNet~\cite{attunet}, VIT~\cite{vit}, R50 VIT~\cite{vit}, TransUNet~\cite{transunet}, Swin-UNet~\cite{swinunet}, LeVit-Unet-384~\cite{levitunet}, nnFormer~\cite{nnformer}, and MISSFormer~\cite{missformer}. 

Quantitative results on the Synapse dataset are reported in Table~\ref{table:synapse}, which show that our method outperforms the previous work by a clear margin. It is notable that the concurrent Transformer-based methods nnFormer and MISSFormer achieve some performance gains compared to the CNN-based methods, while our method still brings further improvement in the average DSC by 1.43\% compared to nnFormer and by 6.87\% compared to MISSFormer. Besides, our D-Former obtains accuracy improvement on almost every organ class, except for the pancreas and stomach, which verifies that our D-Former is a promising and robust framework.

Quantitative results on the ACDC dataset are reported in Table \ref{table:acdc}, and a similar conclusion can be drawn. D-Former achieves the best average DSC of 92.29\% 
without pre-training. Compared with the other methods, our method brings improvements in the average DSC by 4.74\% over R50 U-Net and by 5.54\% over R50 Att-UNet. Compared to the concurrent Transformer-based methods, our method still achieves 0.51\% performance gain over nnFormer and 4.39\% over MISSFormer in the average DSC. Specifically, among all
the key parts of the heart, including the right ventricle (RV), myocardium (Myo), and left ventricle (LV), our D-Former obtains the best segmentation accuracy compared to the other methods in the average DSC. 

From the results in Table~\ref{table:synapse} and Table \ref{table:acdc}, one can see that our D-Former attains excellent generalization on both CT data and MRI data, outperforming the previous methods. Notably, different from most of the known Transformer based frameworks that require a pre-training process, D-Former is initialized randomly and is trained from scratch, yet still obtains competitive performances. This implies that our model could be more suitable for medical imaging tasks when general large size medical image pre-training datasets (such as ImageNet~\cite{imagenet} for natural scene images) are lacking.

\begin{table}
\centering
\caption{Comparison of the numbers of parameters and FLOPs among various methods that segment 3D medical images directly.}
\label{table:flops}
\setlength{\tabcolsep}{7mm}
\begin{threeparttable}
\begin{tabular}{l|cc}
\hline
\multicolumn{1}{c|}{Method}    & \#Params (M) & FLOPs (G) \\ \hline
3D U-Net  & 16.31       & 947.69   \\
UNETR    & 92.25       & 86.02    \\
CoTr     & 41.86       & 377.48   \\
TransBTS & 32.19       & 171.30   \\
nnFormer & 158.92     & 157.88   \\
D-Former & 44.26       & 54.46 \\ \hline
\end{tabular}
\footnotesize               
* The numbers of FLOPs are computed with the input image size of $D\times W \times H=64\times128\times128$. 
\end{threeparttable}
\end{table}

\subsection{Comparison of Model Complexity}
In Table \ref{table:flops}, we compare the numbers of parameters and floating point operations (FLOPs) of our proposed D-Former with those of different 3D medical image segmentation models, including UNETR~\cite{unetr}, CoTr~\cite{cotr}, TransBTS~\cite{transbts}, and nnFormer~\cite{nnformer}. The number of FLOPs is calculated based on the input image size of 64$\times$128$\times$128 for fair comparison. We should note that we omit the part of the complexity brought by activation functions and normalization layers. From Table \ref{table:flops}, one can see that our D-Former has 44.26M parameters and 54.46G FLOPs, which has a lower computational cost compared to nnFormer (157.88G FLOPs), TransBTS (171.30G FLOPs), CoTr (377.48G FLOPs), UNETR (86.02G FLOPs), and 3D U-Net (947.69G FLOPs). Moreover, even if CNN-based models (e.g., 3D U-Net) have less parameters, they are still burdened with high model complexity.

\begin{figure*}
\centering 
\includegraphics[width=1.0\textwidth]{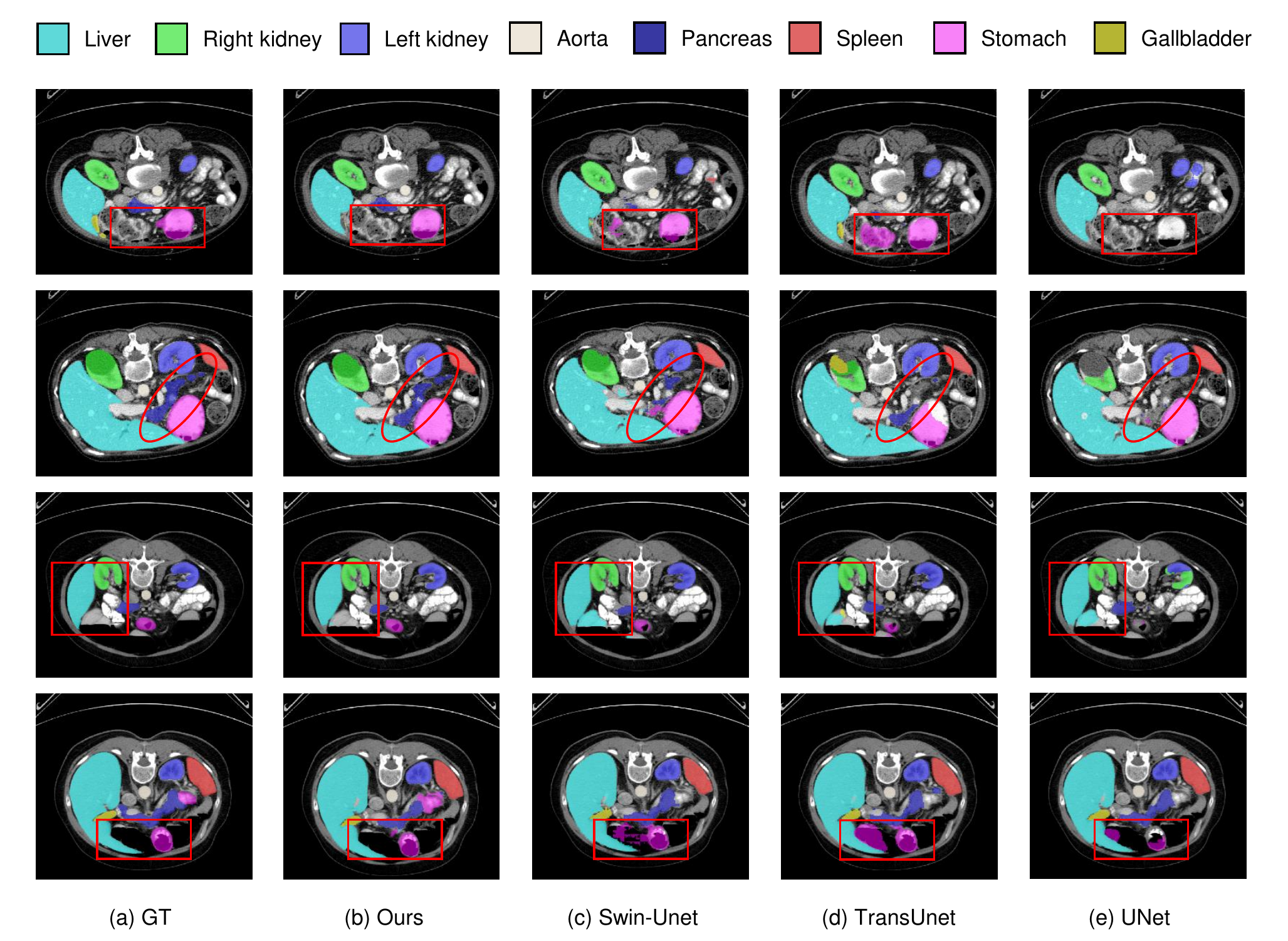}
\caption{Visual comparison with several state-of-the-art methods on the Synapse dataset. The red-marked regions indicate that our model attains discriminative segmentation performance.} 
\label{fig:vis} 
\end{figure*}

\subsection{Qualitative Visualizations}
To intuitively demonstrate the performances of our D-Former model, we compare some qualitative results of our model with several other methods (including Swin-Unet, TransUnet, and UNet) on the Synapse dataset in Fig.~\ref{fig:vis}. One can see that the predicted organ masks of our model are much more similar to the ground truth in general. As for specific organs, our model has better accuracy in identifying and sketching the contours of stomach (e.g., the first and fourth rows), which is consistent with the conclusions based on the quantitative results above. In the second row, only our model can delineate the outline of pancreas well, thus suggesting that our model has a better ability to capture long-range dependency given the fact that the shape of pancreas is long and narrow.  In addition, as illustrated in the third row, our D-Former is able to identify the true region of liver, while the other three models incur some mistakes on the liver. This shows that our method is effective at exploiting the relations between the target organs' patches and the other patches, owing to our model's dynamic position encoding block. In a nutshell, the qualitative visualizations convey an intuitive demonstration of our model's high segmentation accuracy, especially on some slices that are difficult to segment.

\begin{table*}[t]
\centering
\caption{Ablation study on the effect of the Global Scope Module (GSM) (average Dice Similarity Coefficient (DSC) in \%).}
\label{table:nogsm}
\setlength{\tabcolsep}{1mm}
\resizebox{\textwidth}{6mm}{
\begin{tabular}{l|c|cccccccc}
\hline
\multicolumn{1}{c|}{Method}    & Average        & Aotra          & Gallbladder    & Kidnery (L)     & Kidnery (R)     & Liver          & Pancreas       & Spleen         & Stomach        \\ \hline
w/o GSM    & 88.17          & 91.00          & 79.53          & 92.40          & 91.62          & 96.39          & \textbf{77.68} & 92.74          & 84.00          \\
GSM (ours) & \textbf{88.83} & \textbf{92.12} & \textbf{80.09} & \textbf{92.60} & \textbf{91.91} & \textbf{96.99} & 76.67          & \textbf{93.78} & \textbf{86.44}\\ 
\hline
\end{tabular}}
\end{table*}

\begin{table*}[t]
\centering
\caption{Ablation study on the effect of the Global Scope Multi-head Self-attention (GS-MSA) (average Dice Similarity Coefficient (DSC) in \%).}
\label{table:swin}
\setlength{\tabcolsep}{1mm}
\resizebox{\textwidth}{6mm}{
\begin{tabular}{l|c|cccccccc}
\hline
\multicolumn{1}{c|}{Method}       & Average        & Aotra          & Gallbladder    & Kidnery (L)     & Kidnery (R)     & Liver          & Pancreas       & Spleen         & Stomach        \\ \hline
SW-MSA        & 87.50          & 91.37          & 77.48          & 92.55          & \textbf{93.07} & 96.26          & 76.62          & 88.36          & 84.27          \\
GS-MSA (ours) & \textbf{88.83} & \textbf{92.12} & \textbf{80.09} & \textbf{92.60} & 91.91          & \textbf{96.99} & \textbf{76.67} & \textbf{93.78} & \textbf{86.44}\\ 
\hline
\end{tabular}}
\end{table*}

\begin{table*}[t]
\centering
\caption{Ablation study on the effect of dynamic position encoding (DPE) (average Dice Similarity Coefficient (DSC) in \%).}
\label{table:pes}
\setlength{\tabcolsep}{1mm}
\resizebox{\textwidth}{9mm}{
\begin{tabular}{l|c|cccccccc}
\hline
\multicolumn{1}{c|}{Method}       & Average        & Aotra          & Gallbladder    & Kidnery (L)     & Kidnery (R)     & Liver          & Pancreas       & Spleen         & Stomach        \\ \hline
APE        & 84.78          & 88.14          & 76.65          & 90.16          & 86.38          & 95.65          & 67.47          & 91.35          & 82.47          \\
SPE        & 86.04          & 88.51          & 77.13          & \textbf{93.75} & 88.47          & 96.40          & 68.47          & 92.77          & 82.81          \\
RPE        & 86.35          & 90.37          & 78.41          & 92.30          & 87.37          & 96.06          & 70.86          & 91.70          & 83.76          \\
DPE (ours) & \textbf{88.83} & \textbf{92.12} & \textbf{80.09} & 92.60          & \textbf{91.91} & \textbf{96.99} & \textbf{76.67} & \textbf{93.78} & \textbf{86.44}\\
\hline
\end{tabular}}
\end{table*}

\begin{table*}[t]
\centering
\caption{Ablation study on the position of the dynamic position encoding block (average Dice Similarity Coefficient (DSC) in \%).}
\label{table:dpe}
\setlength{\tabcolsep}{1mm}
\resizebox{\textwidth}{7mm}{
\begin{tabular}{l|c|cccccccc}
\hline
\multicolumn{1}{c|}{Method}       & Average        & Aotra          & Gallbladder    & Kidnery (L)     & Kidnery (R)     & Liver          & Pancreas       & Spleen         & Stomach        \\ \hline
After the 1st LSM         & 87.91          & 91.60          & 77.14          & \textbf{93.97} & 90.31          & 95.64          & \textbf{77.39} & 90.64          & \textbf{86.59} \\
After the 1st GSM         & 88.20          & 91.85          & 78.16          & 93.51          & 91.31          & 95.68          & 76.00          & 92.72          & 86.36          \\
Before the 1st LSM (ours) & \textbf{88.83} & \textbf{92.12} & \textbf{80.09} & 92.60          & \textbf{91.91} & \textbf{96.99} & 76.67          & \textbf{93.78} & 86.44       \\  
\hline
\end{tabular}}
\end{table*}

\subsection{Ablation Studies}
We conduct ablation studies on the Synapse dataset to evaluate the effectiveness of our model design. \\

\noindent \textbf{Effect of global scope module (GSM). }
To investigate the necessity of the Global Scope Module (GSM), we replace it by the Local Scope Module (LSM), with the other architectural components unchanged. As shown in Table \ref{table:nogsm}, one can see that the GSM is beneficial to the segmentation accuracy, outperforming using only LSM modules by 0.66\% in the average DSC. This verifies the necessity to explore global interactions of patches across units.\\

\noindent \textbf{Global scope multi-head self-attention (GS-MSA) vs.~other self-attention. }
In order to confirm the effectiveness of our GS-MSA, we compare it with the shift window strategy proposed in Swin Transformer~\cite{swintransformer} which achieves state-of-the-art performance in multiple computer vision tasks. Similar to our GS-MSA design, the shift window strategy (SW-MSA) aims to introduce global attention. Table \ref{table:swin} shows that our global attention design surpasses that in Swin Transformer by 1.33\% in the average DSC. 
\\

\noindent \textbf{Dynamic position encoding vs.~other position encodings.}
We compare our dynamic position encoding (DPE) with other common position encoding methods, including the relative position encoding (RPE)~\cite{rpe,swintransformer}, absolute position encoding (APE)~\cite{transformer}, and sinusoidal position encoding (SPE)~\cite{transformer}. The results are shown in Table~\ref{table:pes}. Compared to APE, SPE, and RPE, our DPE improves them by 4.05\%, 2.79\%, and 2.48\% in the average DSC, respectively.
\\

\begin{wraptable}{r}{6cm}
\centering 
\includegraphics[width=0.4\textwidth]{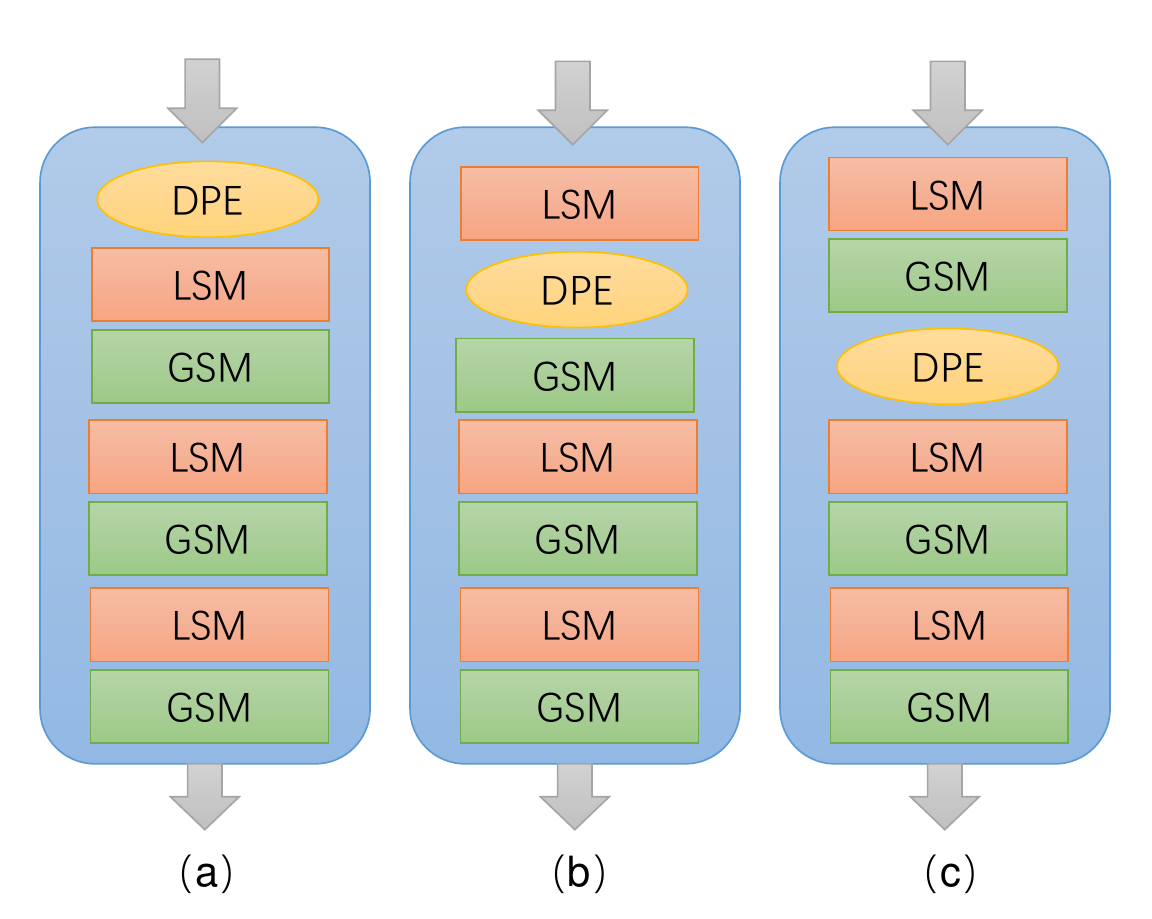}
\caption{Different positions to apply the DPE block. D-Former block 3 is used as an example for illustration, which contains three LSMs and three GSMs, arranging in an alternate manner.} 
\label{fig:dpe} 
\end{wraptable}

\noindent \textbf{Position of the dynamic position encoding block. }
We conduct experiments to examine the performances of different choices of positions to apply the dynamic position encoding block, including placing it (a) before the first LSM, (b) right after the first LSM, and (c) right after the first GSM, in every D-Former block, as illustrated in Fig.~\ref{fig:dpe} taking D-Former block 3 as an example. From Table \ref{table:dpe}, one may find that introducing the position information before the first LSM provides the best segmentation outcomes.

\section{Conclusions}
In this paper, we proposed a novel 3D medical image segmentation framework called D-Former, which utilizes the common U-shape encoder-decoder design and is constructed based on our new dilated Transformer. Our proposed D-Former model can achieve both good efficiency and accuracy, due to its reduced number of patches used in self-attention in local scope module (LSM) and its exploration of long-range dependency with a dilated scope of attention in global scope module (GSM). Moreover, we introduced the dynamic position encoding block, making it possible to flexibly learn vital position information within input sequences. In this way, our model not only reduces the model parameters and decreases the FLOPs, but also attains state-of-the-art semantic segmentation performance on the Synapse and ACDC datasets. 

\backmatter

\bmhead{Acknowledgments}
This research was partially supported by the National Key R\&D Program of China under grant No. 2019YFB1404802, National Natural Science Foundation of China under grants No. 62176231 and 62106218, Zhejiang public welfare technology research project under grant No. LGF20F020013, Wenzhou Bureau of Science and Technology of China under grant  No. Y2020082. D. Z. Chen’s research was supported in part by NSF Grant CCF-1617735.


\begin{thebibliography}{82}
\ifx \bisbn   \undefined \def \bisbn  #1{ISBN #1}\fi
\ifx \binits  \undefined \def \binits#1{#1}\fi
\ifx \bauthor  \undefined \def \bauthor#1{#1}\fi
\ifx \batitle  \undefined \def \batitle#1{#1}\fi
\ifx \bjtitle  \undefined \def \bjtitle#1{#1}\fi
\ifx \bvolume  \undefined \def \bvolume#1{\textbf{#1}}\fi
\ifx \byear  \undefined \def \byear#1{#1}\fi
\ifx \bissue  \undefined \def \bissue#1{#1}\fi
\ifx \bfpage  \undefined \def \bfpage#1{#1}\fi
\ifx \blpage  \undefined \def \blpage #1{#1}\fi
\ifx \burl  \undefined \def \burl#1{\textsf{#1}}\fi
\ifx \doiurl  \undefined \def \doiurl#1{\url{https://doi.org/#1}}\fi
\ifx \betal  \undefined \def \betal{\textit{et al.}}\fi
\ifx \binstitute  \undefined \def \binstitute#1{#1}\fi
\ifx \binstitutionaled  \undefined \def \binstitutionaled#1{#1}\fi
\ifx \bctitle  \undefined \def \bctitle#1{#1}\fi
\ifx \beditor  \undefined \def \beditor#1{#1}\fi
\ifx \bpublisher  \undefined \def \bpublisher#1{#1}\fi
\ifx \bbtitle  \undefined \def \bbtitle#1{#1}\fi
\ifx \bedition  \undefined \def \bedition#1{#1}\fi
\ifx \bseriesno  \undefined \def \bseriesno#1{#1}\fi
\ifx \blocation  \undefined \def \blocation#1{#1}\fi
\ifx \bsertitle  \undefined \def \bsertitle#1{#1}\fi
\ifx \bsnm \undefined \def \bsnm#1{#1}\fi
\ifx \bsuffix \undefined \def \bsuffix#1{#1}\fi
\ifx \bparticle \undefined \def \bparticle#1{#1}\fi
\ifx \barticle \undefined \def \barticle#1{#1}\fi
\bibcommenthead
\ifx \bconfdate \undefined \def \bconfdate #1{#1}\fi
\ifx \botherref \undefined \def \botherref #1{#1}\fi
\ifx \url \undefined \def \url#1{\textsf{#1}}\fi
\ifx \bchapter \undefined \def \bchapter#1{#1}\fi
\ifx \bbook \undefined \def \bbook#1{#1}\fi
\ifx \bcomment \undefined \def \bcomment#1{#1}\fi
\ifx \oauthor \undefined \def \oauthor#1{#1}\fi
\ifx \citeauthoryear \undefined \def \citeauthoryear#1{#1}\fi
\ifx \endbibitem  \undefined \def \endbibitem {}\fi
\ifx \bconflocation  \undefined \def \bconflocation#1{#1}\fi
\ifx \arxivurl  \undefined \def \arxivurl#1{\textsf{#1}}\fi
\csname PreBibitemsHook\endcsname

\bibitem{transunet}
\begin{botherref}
\oauthor{\bsnm{Chen}, \binits{J.}},
\oauthor{\bsnm{Lu}, \binits{Y.}},
\oauthor{\bsnm{Yu}, \binits{Q.}}, et al.:
{TransUNet: Transformers make strong encoders for medical image segmentation}.
ArXiv Preprint ArXiv:2102.04306
(2021)
\end{botherref}
\endbibitem

\bibitem{largekernel}
\begin{bchapter}
\bauthor{\bsnm{Peng}, \binits{C.}},
\bauthor{\bsnm{Zhang}, \binits{X.}},
\bauthor{\bsnm{Yu}, \binits{G.}}, \betal:
\bctitle{{Large kernel matters--{Improve} semantic segmentation by global
  convolutional network}}.
In: \bbtitle{CVPR}
(\byear{2017})
\end{bchapter}
\endbibitem

\bibitem{dilated1}
\begin{botherref}
\oauthor{\bsnm{Yu}, \binits{F.}},
\oauthor{\bsnm{Koltun}, \binits{V.}}:
Multi-scale context aggregation by dilated convolutions.
ArXiv Preprint ArXiv:1511.07122
(2015)
\end{botherref}
\endbibitem

\bibitem{dilated2}
\begin{botherref}
\oauthor{\bsnm{Chen}, \binits{L.-C.}},
\oauthor{\bsnm{Papandreou}, \binits{G.}},
\oauthor{\bsnm{Kokkinos}, \binits{I.}}, et al.:
{DeepLab: Semantic image segmentation with deep convolutional nets, atrous
  convolution, and fully connected CRFs}.
PAMI
(2017)
\end{botherref}
\endbibitem

\bibitem{dilated3}
\begin{bchapter}
\bauthor{\bsnm{Chen}, \binits{L.-C.}},
\bauthor{\bsnm{Zhu}, \binits{Y.}},
\bauthor{\bsnm{Papandreou}, \binits{G.}}, \betal:
\bctitle{Encoder-decoder with atrous separable convolution for semantic image
  segmentation}.
In: \bbtitle{ECCV}
(\byear{2018})
\end{bchapter}
\endbibitem

\bibitem{dilated4}
\begin{botherref}
\oauthor{\bsnm{Gu}, \binits{Z.}},
\oauthor{\bsnm{Cheng}, \binits{J.}},
\oauthor{\bsnm{Fu}, \binits{H.}}, et al.:
{CE-Net: Context encoder network for 2D medical image segmentation}.
TMI
(2019)
\end{botherref}
\endbibitem

\bibitem{pyramid1}
\begin{bchapter}
\bauthor{\bsnm{Zhao}, \binits{H.}},
\bauthor{\bsnm{Shi}, \binits{J.}},
\bauthor{\bsnm{Qi}, \binits{X.}}, \betal:
\bctitle{Pyramid scene parsing network}.
In: \bbtitle{CVPR}
(\byear{2017})
\end{bchapter}
\endbibitem

\bibitem{pyramid2}
\begin{bchapter}
\bauthor{\bsnm{Roth}, \binits{H.R.}},
\bauthor{\bsnm{Shen}, \binits{C.}},
\bauthor{\bsnm{Oda}, \binits{H.}}, \betal:
\bctitle{A multi-scale pyramid of {3D} fully convolutional networks for
  abdominal multi-organ segmentation}.
In: \bbtitle{MICCAI}
(\byear{2018})
\end{bchapter}
\endbibitem

\bibitem{pyramid3}
\begin{botherref}
\oauthor{\bsnm{Feng}, \binits{S.}},
\oauthor{\bsnm{Zhao}, \binits{H.}},
\oauthor{\bsnm{Shi}, \binits{F.}}, et al.:
{CPFNet: Context pyramid fusion network for medical image segmentation}.
TMI
(2020)
\end{botherref}
\endbibitem

\bibitem{deformable1}
\begin{bchapter}
\bauthor{\bsnm{Dai}, \binits{J.}},
\bauthor{\bsnm{Qi}, \binits{H.}},
\bauthor{\bsnm{Xiong}, \binits{Y.}}, \betal:
\bctitle{Deformable convolutional networks}.
In: \bbtitle{ICCV}
(\byear{2017})
\end{bchapter}
\endbibitem

\bibitem{deformable2}
\begin{botherref}
\oauthor{\bsnm{Heinrich}, \binits{M.P.}},
\oauthor{\bsnm{Oktay}, \binits{O.}},
\oauthor{\bsnm{Bouteldja}, \binits{N.}}:
{OBELISK-Net: Fewer layers to solve 3D multi-organ segmentation with sparse
  deformable convolutions}.
MIA
(2019)
\end{botherref}
\endbibitem

\bibitem{deformable3}
\begin{bchapter}
\bauthor{\bsnm{Li}, \binits{Z.}},
\bauthor{\bsnm{Pan}, \binits{H.}},
\bauthor{\bsnm{Zhu}, \binits{Y.}}, \betal:
\bctitle{{PGD-UNet}: A position-guided deformable network for simultaneous
  segmentation of organs and tumors}.
In: \bbtitle{International Joint Conference on Neural Networks}
(\byear{2020})
\end{bchapter}
\endbibitem

\bibitem{detection1}
\begin{bchapter}
\bauthor{\bsnm{Sun}, \binits{Z.}},
\bauthor{\bsnm{Cao}, \binits{S.}},
\bauthor{\bsnm{Yang}, \binits{Y.}}, \betal:
\bctitle{Rethinking {Transformer-based} set prediction for object detection}.
In: \bbtitle{ICCV}
(\byear{2021})
\end{bchapter}
\endbibitem

\bibitem{detection2}
\begin{bchapter}
\bauthor{\bsnm{Pan}, \binits{X.}},
\bauthor{\bsnm{Xia}, \binits{Z.}},
\bauthor{\bsnm{Song}, \binits{S.}}, \betal:
\bctitle{{3D object detection with pointformer}}.
In: \bbtitle{CVPR}
(\byear{2021})
\end{bchapter}
\endbibitem

\bibitem{ant}
\begin{botherref}
\oauthor{\bsnm{Zhang}, \binits{Z.}},
\oauthor{\bsnm{Zhang}, \binits{H.}},
\oauthor{\bsnm{Zhao}, \binits{L.}}, et al.:
Aggregating nested {Transformers}.
ArXiv Preprint ArXiv:2105.12723
(2021)
\end{botherref}
\endbibitem

\bibitem{define}
\begin{botherref}
\oauthor{\bsnm{Mehta}, \binits{S.}},
\oauthor{\bsnm{Koncel-Kedziorski}, \binits{R.}},
\oauthor{\bsnm{Rastegari}, \binits{M.}},
\oauthor{\bsnm{Hajishirzi}, \binits{H.}}:
{DeFINE: DEep Factorized INput Token Embeddings for neural sequence modeling}.
ArXiv Preprint ArXiv:1911.12385
(2020)
\end{botherref}
\endbibitem

\bibitem{delight}
\begin{botherref}
\oauthor{\bsnm{Mehta}, \binits{S.}},
\oauthor{\bsnm{Ghazvininejad}, \binits{M.}},
\oauthor{\bsnm{Iyer}, \binits{S.}}, et al.:
{DeLighT}: Very deep and light-weight {Transformer}.
CoRR
(2020)
\end{botherref}
\endbibitem

\bibitem{fcnn}
\begin{bchapter}
\bauthor{\bsnm{Long}, \binits{J.}},
\bauthor{\bsnm{Shelhamer}, \binits{E.}},
\bauthor{\bsnm{Darrell}, \binits{T.}}:
\bctitle{Fully convolutional networks for semantic segmentation}.
In: \bbtitle{CVPR}
(\byear{2015})
\end{bchapter}
\endbibitem

\bibitem{unet3+}
\begin{bchapter}
\bauthor{\bsnm{Huang}, \binits{H.}},
\bauthor{\bsnm{Lin}, \binits{L.}},
\bauthor{\bsnm{Tong}, \binits{R.}}, \betal:
\bctitle{{UNet} 3+: A full-scale connected {UNet} for medical image
  segmentation}.
In: \bbtitle{IEEE International Conference on Acoustics, Speech and Signal
  Processing}
(\byear{2020})
\end{bchapter}
\endbibitem

\bibitem{unet}
\begin{bchapter}
\bauthor{\bsnm{Ronneberger}, \binits{O.}},
\bauthor{\bsnm{Fischer}, \binits{P.}},
\bauthor{\bsnm{Brox}, \binits{T.}}:
\bctitle{{U-Net}: Convolutional networks for biomedical image segmentation}.
In: \bbtitle{MICCAI}
(\byear{2015})
\end{bchapter}
\endbibitem

\bibitem{polylr}
\begin{bchapter}
\bauthor{\bsnm{Mishra}, \binits{P.}},
\bauthor{\bsnm{Sarawadekar}, \binits{K.}}:
\bctitle{Polynomial learning rate policy with warm restart for deep neural
  network}.
In: \bbtitle{IEEE Region 10 Conference}
(\byear{2019})
\end{bchapter}
\endbibitem

\bibitem{celoss}
\begin{bchapter}
\bauthor{\bsnm{Yi-de}, \binits{M.}},
\bauthor{\bsnm{Qing}, \binits{L.}},
\bauthor{\bsnm{Zhi-Bai}, \binits{Q.}}:
\bctitle{Automated image segmentation using improved {PCNN} model based on
  cross-entropy}.
In: \bbtitle{International Symposium on Intelligent Multimedia, Video and
  Speech Processing}
(\byear{2004})
\end{bchapter}
\endbibitem

\bibitem{losssurvey}
\begin{bchapter}
\bauthor{\bsnm{Jadon}, \binits{S.}}:
\bctitle{A survey of loss functions for semantic segmentation}.
In: \bbtitle{IEEE Conference on Computational Intelligence in Bioinformatics
  and Computational Biology}
(\byear{2020})
\end{bchapter}
\endbibitem

\bibitem{cpe}
\begin{botherref}
\oauthor{\bsnm{Chu}, \binits{X.}},
\oauthor{\bsnm{Tian}, \binits{Z.}},
\oauthor{\bsnm{Zhang}, \binits{B.}}, et al.:
Conditional positional encodings for vision {Transformers}.
ArXiv Preprint ArXiv:2102.10882
(2021)
\end{botherref}
\endbibitem

\bibitem{transinvar}
\begin{botherref}
\oauthor{\bsnm{Kauderer-Abrams}, \binits{E.}}:
Quantifying translation-invariance in convolutional neural networks.
ArXiv Preprint ArXiv:1801.01450
(2017)
\end{botherref}
\endbibitem

\bibitem{darr}
\begin{bchapter}
\bauthor{\bsnm{Fu}, \binits{S.}},
\bauthor{\bsnm{Lu}, \binits{Y.}},
\bauthor{\bsnm{Wang}, \binits{Y.}}, \betal:
\bctitle{Domain adaptive relational reasoning for {3D} multi-organ
  segmentation}.
In: \bbtitle{MICCAI}
(\byear{2020})
\end{bchapter}
\endbibitem

\bibitem{transbts}
\begin{bchapter}
\bauthor{\bsnm{Wang}, \binits{W.}},
\bauthor{\bsnm{Chen}, \binits{C.}},
\bauthor{\bsnm{Ding}, \binits{M.}}, \betal:
\bctitle{{TransBTS}: Multimodal brain tumor segmentation using {Transformer}}.
In: \bbtitle{MICCAI}
(\byear{2021})
\end{bchapter}
\endbibitem

\bibitem{levitunet}
\begin{botherref}
\oauthor{\bsnm{Xu}, \binits{G.}},
\oauthor{\bsnm{Wu}, \binits{X.}},
\oauthor{\bsnm{Zhang}, \binits{X.}}, et al.:
{LeViT-UNet}: Make faster encoders with {Transformer} for medical image
  segmentation.
ArXiv Preprint ArXiv:2107.08623
(2021)
\end{botherref}
\endbibitem

\bibitem{unet++}
\begin{bchapter}
\bauthor{\bsnm{Zhou}, \binits{Z.}},
\bauthor{\bsnm{Siddiquee}, \binits{M.M.R.}},
\bauthor{\bsnm{Tajbakhsh}, \binits{N.}}, \betal:
\bctitle{{UNet++}: A nested {U-Net} architecture for medical image
  segmentation}.
In: \bbtitle{Deep Learning in Medical Image Analysis and Multimodal Learning
  for Clinical Decision Support},
(\byear{2018})
\end{bchapter}
\endbibitem

\bibitem{raunet}
\begin{bchapter}
\bauthor{\bsnm{Ni}, \binits{Z.-L.}},
\bauthor{\bsnm{Bian}, \binits{G.-B.}},
\bauthor{\bsnm{Zhou}, \binits{X.-H.}}, \betal:
\bctitle{{RAUNet}: Residual attention u-net for semantic segmentation of
  cataract surgical instruments}.
In: \bbtitle{International Conference on Neural Information Processing}
(\byear{2019})
\end{bchapter}
\endbibitem

\bibitem{3dunet}
\begin{bchapter}
\bauthor{\bsnm{{\c{C}}i{\c{c}}ek}, \binits{{\"O}.}},
\bauthor{\bsnm{Abdulkadir}, \binits{A.}},
\bauthor{\bsnm{Lienkamp}, \binits{S.S.}}, \betal:
\bctitle{{3D U-Net}: Learning dense volumetric segmentation from sparse
  annotation}.
In: \bbtitle{MICCAI}
(\byear{2016})
\end{bchapter}
\endbibitem

\bibitem{nnunet}
\begin{botherref}
\oauthor{\bsnm{Isensee}, \binits{F.}},
\oauthor{\bsnm{Jaeger}, \binits{P.F.}},
\oauthor{\bsnm{Kohl}, \binits{S.A.}}, et al.:
{nnU-Net: A self-configuring method for deep learning-based biomedical image
  segmentation}.
Nature Methods
(2021)
\end{botherref}
\endbibitem

\bibitem{resunet}
\begin{botherref}
\oauthor{\bsnm{Diakogiannis}, \binits{F.I.}},
\oauthor{\bsnm{Waldner}, \binits{F.}},
\oauthor{\bsnm{Caccetta}, \binits{P.}}, et al.:
{ResUNet-a: A deep learning framework for semantic segmentation of remotely
  sensed data}.
Journal of Photogrammetry and Remote Sensing
(2020)
\end{botherref}
\endbibitem

\bibitem{transformer}
\begin{bchapter}
\bauthor{\bsnm{Vaswani}, \binits{A.}},
\bauthor{\bsnm{Shazeer}, \binits{N.}},
\bauthor{\bsnm{Parmar}, \binits{N.}}, \betal:
\bctitle{Attention is all you need}.
In: \bbtitle{NIPS}
(\byear{2017})
\end{bchapter}
\endbibitem

\bibitem{vit}
\begin{botherref}
\oauthor{\bsnm{Dosovitskiy}, \binits{A.}},
\oauthor{\bsnm{Beyer}, \binits{L.}},
\oauthor{\bsnm{Kolesnikov}, \binits{A.}}, et al.:
An image is worth 16x16 words: {Transformers} for image recognition at scale.
ArXiv Preprint ArXiv:2010.11929
(2020)
\end{botherref}
\endbibitem

\bibitem{bert}
\begin{botherref}
\oauthor{\bsnm{Devlin}, \binits{J.}},
\oauthor{\bsnm{Chang}, \binits{M.-W.}},
\oauthor{\bsnm{Lee}, \binits{K.}}, et al.:
Bert: Pre-training of deep bidirectional {Transformers} for language
  understanding.
ArXiv Preprint ArXiv:1810.04805
(2018)
\end{botherref}
\endbibitem

\bibitem{detr}
\begin{bchapter}
\bauthor{\bsnm{Carion}, \binits{N.}},
\bauthor{\bsnm{Massa}, \binits{F.}},
\bauthor{\bsnm{Synnaeve}, \binits{G.}}, \betal:
\bctitle{End-to-end object detection with {Transformers}}.
In: \bbtitle{ECCV}
(\byear{2020})
\end{bchapter}
\endbibitem

\bibitem{setr}
\begin{bchapter}
\bauthor{\bsnm{Zheng}, \binits{S.}},
\bauthor{\bsnm{Lu}, \binits{J.}},
\bauthor{\bsnm{Zhao}, \binits{H.}}, \betal:
\bctitle{Rethinking semantic segmentation from a sequence-to-sequence
  perspective with {Transformers}}.
In: \bbtitle{CVPR}
(\byear{2021})
\end{bchapter}
\endbibitem

\bibitem{deformabledetr}
\begin{botherref}
\oauthor{\bsnm{Zhu}, \binits{X.}},
\oauthor{\bsnm{Su}, \binits{W.}},
\oauthor{\bsnm{Lu}, \binits{L.}}, et al.:
Deformable {DETR}: Deformable {Transformers} for end-to-end object detection.
ArXiv Preprint ArXiv:2010.04159
(2020)
\end{botherref}
\endbibitem

\bibitem{deit}
\begin{bchapter}
\bauthor{\bsnm{Touvron}, \binits{H.}},
\bauthor{\bsnm{Cord}, \binits{M.}},
\bauthor{\bsnm{Douze}, \binits{M.}}, \betal:
\bctitle{Training data-efficient image {Transformers} \& distillation through
  attention}.
In: \bbtitle{ICML}
(\byear{2021})
\end{bchapter}
\endbibitem

\bibitem{nonlocal}
\begin{bchapter}
\bauthor{\bsnm{Wang}, \binits{X.}},
\bauthor{\bsnm{Girshick}, \binits{R.}},
\bauthor{\bsnm{Gupta}, \binits{A.}}, \betal:
\bctitle{Non-local neural networks}.
In: \bbtitle{CVPR}
(\byear{2018})
\end{bchapter}
\endbibitem

\bibitem{nnformer}
\begin{botherref}
\oauthor{\bsnm{Zhou}, \binits{H.-Y.}},
\oauthor{\bsnm{Guo}, \binits{J.}},
\oauthor{\bsnm{Zhang}, \binits{Y.}}, et al.:
{nnFormer}: Interleaved {Transformers} for volumetric segmentation.
ArXiv Preprint ArXiv:2109.03201
(2021)
\end{botherref}
\endbibitem

\bibitem{missformer}
\begin{botherref}
\oauthor{\bsnm{Huang}, \binits{X.}},
\oauthor{\bsnm{Deng}, \binits{Z.}},
\oauthor{\bsnm{Li}, \binits{D.}}, et al.:
{MISSFormer}: An effective medical image segmentation {Transformer}.
ArXiv Preprint ArXiv:2109.07162
(2021)
\end{botherref}
\endbibitem

\bibitem{dstransunet}
\begin{botherref}
\oauthor{\bsnm{Lin}, \binits{A.}},
\oauthor{\bsnm{Chen}, \binits{B.}},
\oauthor{\bsnm{Xu}, \binits{J.}}, et al.:
{DS-TransUNet}: Dual swin {Transformer} {U-Net} for medical image segmentation.
ArXiv Preprint ArXiv:2106.06716
(2021)
\end{botherref}
\endbibitem

\bibitem{pvt}
\begin{botherref}
\oauthor{\bsnm{Wang}, \binits{W.}},
\oauthor{\bsnm{Xie}, \binits{E.}},
\oauthor{\bsnm{Li}, \binits{X.}}, et al.:
Pyramid vision {Transformers}: A versatile backbone for dense prediction
  without convolutions.
ArXiv Preprint ArXiv:2102.12122
(2021)
\end{botherref}
\endbibitem

\bibitem{swintransformer}
\begin{botherref}
\oauthor{\bsnm{Liu}, \binits{Z.}},
\oauthor{\bsnm{Lin}, \binits{Y.}},
\oauthor{\bsnm{Cao}, \binits{Y.}}, et al.:
Swin {Transformers}: Hierarchical vision {Transformers} using shifted windows.
ArXiv Preprint ArXiv:2103.14030
(2021)
\end{botherref}
\endbibitem

\bibitem{t2tvit}
\begin{botherref}
\oauthor{\bsnm{Yuan}, \binits{L.}},
\oauthor{\bsnm{Chen}, \binits{Y.}},
\oauthor{\bsnm{Wang}, \binits{T.}}, et al.:
{Tokens-to-Token ViT}: Training vision {Transformers} from scratch on
  {ImageNet}.
ArXiv Preprint ArXiv:2101.11986
(2021)
\end{botherref}
\endbibitem

\bibitem{volo}
\begin{botherref}
\oauthor{\bsnm{Yuan}, \binits{L.}},
\oauthor{\bsnm{Hou}, \binits{Q.}},
\oauthor{\bsnm{Jiang}, \binits{Z.}}, et al.:
{VOLO}: Vision outlooker for visual recognition.
ArXiv Preprint ArXiv:2106.13112
(2021)
\end{botherref}
\endbibitem

\bibitem{unetr}
\begin{botherref}
\oauthor{\bsnm{Hatamizadeh}, \binits{A.}},
\oauthor{\bsnm{Tang}, \binits{Y.}},
\oauthor{\bsnm{Nath}, \binits{V.}}, et al.:
{UNETR}: {Transformers} for {3D} medical image segmentation.
ArXiv Preprint ArXiv:2103.10504
(2021)
\end{botherref}
\endbibitem

\bibitem{cotr}
\begin{botherref}
\oauthor{\bsnm{Xie}, \binits{Y.}},
\oauthor{\bsnm{Zhang}, \binits{J.}},
\oauthor{\bsnm{Shen}, \binits{C.}}, et al.:
{CoTr}: Efficiently bridging {CNN} and {Transformer} for {3D} medical image
  segmentation.
ArXiv Preprint ArXiv:2103.03024
(2021)
\end{botherref}
\endbibitem

\bibitem{transfuse}
\begin{botherref}
\oauthor{\bsnm{Zhang}, \binits{Y.}},
\oauthor{\bsnm{Liu}, \binits{H.}},
\oauthor{\bsnm{Hu}, \binits{Q.}}:
{TransFuse}: {Fusing Transformers and CNNs} for medical image segmentation.
ArXiv Preprint ArXiv:2102.08005
(2021)
\end{botherref}
\endbibitem

\bibitem{xcit}
\begin{botherref}
\oauthor{\bsnm{El-Nouby}, \binits{A.}},
\oauthor{\bsnm{Touvron}, \binits{H.}},
\oauthor{\bsnm{Caron}, \binits{M.}}, et al.:
{XCiT: Cross-covariance image Transformers}.
ArXiv Preprint ArXiv:2106.09681
(2021)
\end{botherref}
\endbibitem

\bibitem{swinunet}
\begin{botherref}
\oauthor{\bsnm{Cao}, \binits{H.}},
\oauthor{\bsnm{Wang}, \binits{Y.}},
\oauthor{\bsnm{Chen}, \binits{J.}}, et al.:
{Swin-Unet: Unet-like pure Transformer for medical image segmentation}.
ArXiv Preprint ArXiv:2105.05537
(2021)
\end{botherref}
\endbibitem

\bibitem{litetransformer}
\begin{botherref}
\oauthor{\bsnm{Wu}, \binits{Z.}},
\oauthor{\bsnm{Liu}, \binits{Z.}}, et al.:
Lite {Transformer} with long-short range attention.
ArXiv Preprint ArXiv:2004.11886
(2020)
\end{botherref}
\endbibitem

\bibitem{fcn1}
\begin{bchapter}
\bauthor{\bsnm{Korez}, \binits{R.}},
\bauthor{\bsnm{Likar}, \binits{B.}},
\bauthor{\bsnm{Pernu{\v{s}}}, \binits{F.}}, \betal:
\bctitle{Model-based segmentation of vertebral bodies from {MR} images with
  {3D} {CNNs}}.
In: \bbtitle{MICCAI}
(\byear{2016})
\end{bchapter}
\endbibitem

\bibitem{fcn2}
\begin{bchapter}
\bauthor{\bsnm{Zhou}, \binits{X.}},
\bauthor{\bsnm{Ito}, \binits{T.}},
\bauthor{\bsnm{Takayama}, \binits{R.}}, \betal:
\bctitle{Three-dimensional {CT} image segmentation by combining {2D} fully
  convolutional network with {3D} majority voting}.
In: \bbtitle{Deep Learning and Data Labeling for Medical Applications},
(\byear{2016})
\end{bchapter}
\endbibitem

\bibitem{fcn3}
\begin{bchapter}
\bauthor{\bsnm{Moeskops}, \binits{P.}},
\bauthor{\bsnm{Wolterink}, \binits{J.M.}}, \betal:
\bctitle{Deep learning for multi-task medical image segmentation in multiple
  modalities}.
In: \bbtitle{MICCAI}
(\byear{2016})
\end{bchapter}
\endbibitem

\bibitem{fcn4}
\begin{bchapter}
\bauthor{\bsnm{Shakeri}, \binits{M.}},
\bauthor{\bsnm{Tsogkas}, \binits{S.}},
\bauthor{\bsnm{Ferrante}, \binits{E.}}, \betal:
\bctitle{Sub-cortical brain structure segmentation using {F-CNN's}}.
In: \bbtitle{International Symposium on Biomedical Imaging}
(\byear{2016})
\end{bchapter}
\endbibitem

\bibitem{fcn5}
\begin{bchapter}
\bauthor{\bsnm{Alansary}, \binits{A.}},
\bauthor{\bsnm{Kamnitsas}, \binits{K.}},
\bauthor{\bsnm{Davidson}, \binits{A.}}, \betal:
\bctitle{Fast fully automatic segmentation of the human placenta from motion
  corrupted {MRI}}.
In: \bbtitle{MICCAI}
(\byear{2016})
\end{bchapter}
\endbibitem

\bibitem{unet1}
\begin{botherref}
\oauthor{\bsnm{Wang}, \binits{C.}},
\oauthor{\bsnm{MacGillivray}, \binits{T.}},
\oauthor{\bsnm{Macnaught}, \binits{G.}}, et al.:
A two-stage {3D Unet} framework for multi-class segmentation on full resolution
  image.
ArXiv Preprint ArXiv:1804.04341
(2018)
\end{botherref}
\endbibitem

\bibitem{unet2}
\begin{bchapter}
\bauthor{\bsnm{{\c{C}}i{\c{c}}ek}, \binits{{\"O}.}},
\bauthor{\bsnm{Abdulkadir}, \binits{A.}},
\bauthor{\bsnm{Lienkamp}, \binits{S.S.}}, \betal:
\bctitle{{3D U-Net: Learning dense volumetric segmentation from sparse
  annotation}}.
In: \bbtitle{MICCAI}
(\byear{2016})
\end{bchapter}
\endbibitem

\bibitem{unet3}
\begin{botherref}
\oauthor{\bsnm{Kamnitsas}, \binits{K.}},
\oauthor{\bsnm{Ledig}, \binits{C.}},
\oauthor{\bsnm{Newcombe}, \binits{V.F.}}, et al.:
Efficient multi-scale {3D CNN with fully connected CRF for accurate brain
  lesion segmentation}.
MIA
(2017)
\end{botherref}
\endbibitem

\bibitem{unet4}
\begin{bchapter}
\bauthor{\bsnm{Drozdzal}, \binits{M.}},
\bauthor{\bsnm{Vorontsov}, \binits{E.}},
\bauthor{\bsnm{Chartrand}, \binits{G.}}, \betal:
\bctitle{The importance of skip connections in biomedical image segmentation}.
In: \bbtitle{Deep Learning and Data Labeling for Medical Applications},
(\byear{2016})
\end{bchapter}
\endbibitem

\bibitem{unet5}
\begin{bchapter}
\bauthor{\bsnm{Ghafoorian}, \binits{M.}},
\bauthor{\bsnm{Karssemeijer}, \binits{N.}},
\bauthor{\bsnm{Heskes}, \binits{T.}}, \betal:
\bctitle{Non-uniform patch sampling with deep convolutional neural networks for
  white matter hyperintensity segmentation}.
In: \bbtitle{International Symposium on Biomedical Imaging}
(\byear{2016})
\end{bchapter}
\endbibitem

\bibitem{unet6}
\begin{botherref}
\oauthor{\bsnm{Brosch}, \binits{T.}},
\oauthor{\bsnm{Tang}, \binits{L.Y.}},
\oauthor{\bsnm{Yoo}, \binits{Y.}}, et al.:
Deep {3D} convolutional encoder networks with shortcuts for multiscale feature
  integration applied to multiple sclerosis lesion segmentation.
TMI
(2016)
\end{botherref}
\endbibitem

\bibitem{autoliverseg}
\begin{botherref}
\oauthor{\bsnm{Christ}, \binits{P.F.}},
\oauthor{\bsnm{Ettlinger}, \binits{F.}}, et al.:
Automatic liver and tumor segmentation of {CT} and {MRI} volumes using cascaded
  fully convolutional neural networks.
ArXiv Preprint ArXiv:1702.05970
(2017)
\end{botherref}
\endbibitem

\bibitem{lesionseg}
\begin{botherref}
\oauthor{\bsnm{Brosch}, \binits{T.}},
\oauthor{\bsnm{Tang}, \binits{L.Y.}},
\oauthor{\bsnm{Yoo}, \binits{Y.}}, et al.:
Deep {3D} convolutional encoder networks with shortcuts for multiscale feature
  integration applied to multiple sclerosis lesion segmentation.
TMI
(2016)
\end{botherref}
\endbibitem

\bibitem{brainseg}
\begin{botherref}
\oauthor{\bsnm{Pereira}, \binits{S.}},
\oauthor{\bsnm{Pinto}, \binits{A.}}, et al.:
Brain tumor segmentation using convolutional neural networks in {MRI} images.
TMI
(2016)
\end{botherref}
\endbibitem

\bibitem{rpe}
\begin{botherref}
\oauthor{\bsnm{Shaw}, \binits{P.}},
\oauthor{\bsnm{Uszkoreit}, \binits{J.}},
\oauthor{\bsnm{Vaswani}, \binits{A.}}:
Self-attention with relative position representations.
ArXiv Preprint ArXiv:1803.02155
(2018)
\end{botherref}
\endbibitem

\bibitem{denseunet}
\begin{botherref}
\oauthor{\bsnm{Cai}, \binits{S.}},
\oauthor{\bsnm{Tian}, \binits{Y.}},
\oauthor{\bsnm{Lui}, \binits{H.}}, et al.:
{Dense-UNet: A novel multiphoton in vivo cellular image segmentation model
  based on a convolutional neural network}.
Quantitative Imaging in Medicine and Surgery
(2020)
\end{botherref}
\endbibitem

\bibitem{tint}
\begin{botherref}
\oauthor{\bsnm{Han}, \binits{K.}},
\oauthor{\bsnm{Xiao}, \binits{A.}},
\oauthor{\bsnm{Wu}, \binits{E.}}, et al.:
{Transformer in Transformer}.
ArXiv Preprint ArXiv:2103.00112
(2021)
\end{botherref}
\endbibitem

\bibitem{medt}
\begin{botherref}
\oauthor{\bsnm{Valanarasu}, \binits{J.M.J.}},
\oauthor{\bsnm{Oza}, \binits{P.}}, et al.:
Medical {Transformer}: Gated axial-attention for medical image segmentation.
ArXiv Preprint ArXiv:2102.10662
(2021)
\end{botherref}
\endbibitem

\bibitem{deeplab0}
\begin{botherref}
\oauthor{\bsnm{Chen}, \binits{L.-C.}},
\oauthor{\bsnm{Papandreou}, \binits{G.}},
\oauthor{\bsnm{Kokkinos}, \binits{I.}}, et al.:
Semantic image segmentation with deep convolutional nets and fully connected
  {CRFs}.
ArXiv Preprint ArXiv:1412.7062
(2014)
\end{botherref}
\endbibitem

\bibitem{deeplab}
\begin{botherref}
\oauthor{\bsnm{Chen}, \binits{L.-C.}},
\oauthor{\bsnm{Papandreou}, \binits{G.}},
\oauthor{\bsnm{Kokkinos}, \binits{I.}}, et al.:
{DeepLab: Semantic image segmentation with deep convolutional nets, atrous
  convolution, and fully connected CRFs}.
TPAMI
(2017)
\end{botherref}
\endbibitem

\bibitem{deeplab2}
\begin{botherref}
\oauthor{\bsnm{Chen}, \binits{L.-C.}},
\oauthor{\bsnm{Papandreou}, \binits{G.}},
\oauthor{\bsnm{Schroff}, \binits{F.}}, et al.:
Rethinking atrous convolution for semantic image segmentation.
ArXiv Preprint ArXiv:1706.05587
(2017)
\end{botherref}
\endbibitem

\bibitem{deeplab3}
\begin{bchapter}
\bauthor{\bsnm{Chen}, \binits{L.-C.}},
\bauthor{\bsnm{Zhu}, \binits{Y.}},
\bauthor{\bsnm{Papandreou}, \binits{G.}}, \betal:
\bctitle{Encoder-decoder with atrous separable convolution for semantic image
  segmentation}.
In: \bbtitle{ECCV}
(\byear{2018})
\end{bchapter}
\endbibitem

\bibitem{vnet}
\begin{bchapter}
\bauthor{\bsnm{Milletari}, \binits{F.}},
\bauthor{\bsnm{Navab}, \binits{N.}},
\bauthor{\bsnm{Ahmadi}, \binits{S.-A.}}:
\bctitle{{V-Net}: Fully convolutional neural networks for volumetric medical
  image segmentation}.
In: \bbtitle{3DV}
(\byear{2016})
\end{bchapter}
\endbibitem

\bibitem{attunet}
\begin{botherref}
\oauthor{\bsnm{Schlemper}, \binits{J.}},
\oauthor{\bsnm{Oktay}, \binits{O.}},
\oauthor{\bsnm{Schaap}, \binits{M.}}, et al.:
Attention gated networks: Learning to leverage salient regions in medical
  images.
MIA
(2019)
\end{botherref}
\endbibitem

\bibitem{ln}
\begin{botherref}
\oauthor{\bsnm{Ba}, \binits{J.L.}},
\oauthor{\bsnm{Kiros}, \binits{J.R.}},
\oauthor{\bsnm{Hinton}, \binits{G.E.}}:
Layer normalization.
ArXiv Preprint ArXiv:1607.06450
(2016)
\end{botherref}
\endbibitem

\bibitem{dwconv}
\begin{bchapter}
\bauthor{\bsnm{Chollet}, \binits{F.}}:
\bctitle{Xception: Deep learning with depthwise separable convolutions}.
In: \bbtitle{CVPR},
pp. \bfpage{1251}--\blpage{1258}
(\byear{2017})
\end{bchapter}
\endbibitem

\bibitem{imagenet}
\begin{bchapter}
\bauthor{\bsnm{Deng}, \binits{J.}},
\bauthor{\bsnm{Dong}, \binits{W.}},
\bauthor{\bsnm{Socher}, \binits{R.}}, \betal:
\bctitle{{ImageNet}: A large-scale hierarchical image database}.
In: \bbtitle{CVPR}
(\byear{2009})
\end{bchapter}
\endbibitem

\bibitem{sgd}
\begin{bchapter}
\bauthor{\bsnm{Bottou}, \binits{L.}}:
\bctitle{Stochastic gradient descent tricks}.
In: \bbtitle{Neural Networks: Tricks of the Trade},
(\byear{2012})
\end{bchapter}
\endbibitem

\end{thebibliography}
\end{document}